\documentclass{article}

\usepackage{graphicx}
\usepackage{arxiv}
\usepackage[utf8]{inputenc} 
\usepackage[T1]{fontenc}    
\usepackage{hyperref}       
\usepackage{url}            
\usepackage{booktabs}       
\usepackage{amsfonts}       
\usepackage{nicefrac}       
\usepackage{microtype}      
\usepackage{lipsum}
\usepackage{amsmath}
\usepackage{algorithm}
\usepackage{algorithmic}
\usepackage{bbm}
\usepackage{comment}
\usepackage{trace}
\title{Watch It Twice: Video Captioning with a Refocused Video Encoder}

\author{
  Xiangxi Shi\\
  Nanyang Technological University\\
  \texttt{xxshi@ntu.edu.sg} \\
  \And
  Jianfei Cai\\
  Nanyang Technological University\\
  \texttt{asjfcai@ntu.edu.sg}\\
  \And
  Shafiq Joty\\
  Nanyang Technological University\\
  \texttt{srjoty@ntu.edu.sg}\\
  \And
  Jiuxiang Gu\\
  Nanyang Technological University\\
  \texttt{JGU004@e.ntu.edu.sg} \\
}

\begin{document}
\maketitle

\begin{abstract}
With the rapid growth of video data and the increasing demands of various applications such as intelligent video search and assistance toward visually-impaired people, video captioning task has received a lot of attention recently in computer vision and natural language processing fields. The state-of-the-art video captioning methods focus more on encoding the temporal information, while lack of effective ways to remove irrelevant temporal information and also neglecting the spatial details. However, the current RNN encoding module in single time order can be influenced by the irrelevant temporal information, especially the irrelevant temporal information is at the beginning of the encoding. In addition, neglecting spatial information will lead to the relationship confusion of the words and detailed loss. Therefore, in this paper, we propose a novel recurrent video encoding method and a novel visual spatial feature for the video captioning task. The recurrent encoding module encodes the video twice with the predicted key frame to avoid the irrelevant temporal information often occurring at the beginning and the end of a video.
The novel spatial features represent the spatial information in different regions of a video and enrich the details of a caption. Experiments on two benchmark datasets show superior performance of the proposed method.
\end{abstract}

\keywords{video captioning, recurrent video encoding, reinforcement learning, key frame}

\section{Introduction}
\begin{figure*}[ht]
	\centering
	\vspace{2mm}
	\includegraphics[width=\linewidth]{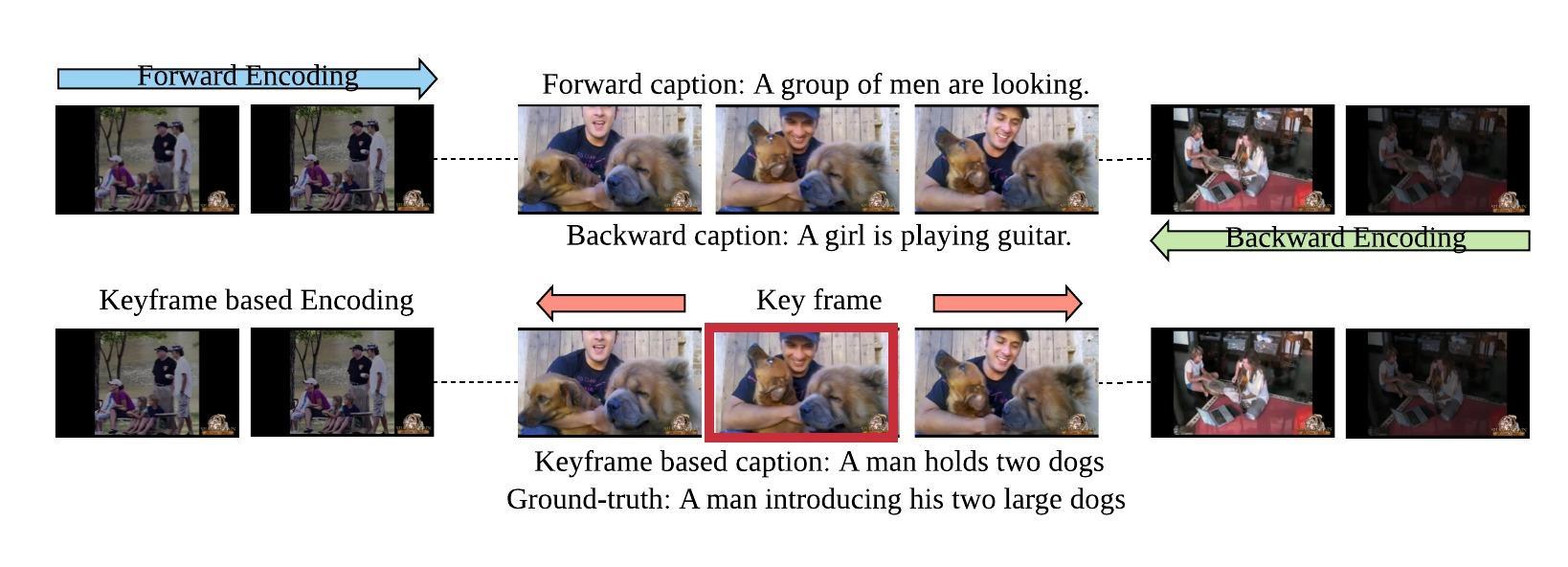} 
	\vspace{-0.3in}
	\caption{Overall idea of the proposed video refocusing encoder. Top: video captioning with forward and backward video encoding. Bottom: the propose key frame based video encoding.}
	\label{fig:intro}
\end{figure*}

Video captioning task aims to automatically generate natural and meaningful textual descriptions for videos. With the explosion of mass video data generated through various electronic devices, automatic video captioning has become a popular task. Video captioning can be used in various applications to make our lives convenient such as generating captions as additional information for intelligent video search and serving as a tool to help visually-impaired people.

The pre-trained convolutional neural network (CNN) have been applied in video captioning tasks successfully to extracting useful frame features \cite{simonyan2014very, he2016deep, szegedy2017inception}. This helps to save memory and training time, and makes the optimization problem easier. However, different tasks require different strategies to pre-process the extracted features. For example, in the image captioning task~\cite{lu2017knowing, anderson2018bottom, yao2018exploring, gu2018stack}, a captioning model is required to generate a caption containing multiple objects in different regions of an image. Thus, spatial information is important for the image captioning task. 

Different from image captioning, in the video captioning task, temporal information is more significant since the actions, event dynamics, and the surroundings in a video can hardly be fully represented by a single image. Recent works~\cite{ABCDE,song2017hierarchical, shi2018video, bin2016bidirectional, chen2018less} 

mainly focus on encoding the temporal relationship in a video to predict a correct sentence, where the spatial feature of a frame is typically aggregated or pooled into one feature vector representing the frame; it is also computationally prohibitive to keep all the spatial features for all the sampled frames.

Then, a Recurrent Neural Network (RNN) is used as a video encoder to compose the sequence of the frame features into a video representation, followed by another RNN as a language decoder to generate the words. 

Such a sequence-to-sequence encoder-decoder architecture is the dominant backbone structure of most of the current video captioning methods. 
    
Despite the great success, the sequence-to-sequence based video captioning methods~\cite{venugopalan2015sequence,bin2016bidirectional,baraldi2017hierarchical} still suffer from some key limitations. First, existing methods assume that a video only narrates a single story and has few scenery changes, and thus they usually apply an RNN encoder to encode a video along the forward direction (from start to end), or the backward direction, or both. Because of the sequential RNN structure, the current input signal can be modified and declined according to the previous hidden state. Thus, in the presence of noisy information at the beginning of a sequence, it will have negative influence on encoding the key information that appear after the noise. In many practical situations, a video often contains many different scenarios and some noisy information that are irrelevant to the main theme of the video and therefore should be disregarded in the caption generation process. As shown in Fig~\ref{fig:intro}, the key event matching the ground truth caption is located at the middle of the video. At the start and at the end of the video, there are some irrelevant frames which can interfere the video dependency and thus lead to inaccurate caption prediction. This motivates us to come up with the novel idea: \textit{how about watching a video twice: first predict the key frame and then encode the video based on that?}
    
Another key problem of the sequence-to-sequence based video captioning methods is that they disregard spatial features of the frames. Temporal information can help predict accurate actions for the sentences, however, we argue that it is often insufficient to represent entity-wise details in different regions of a single frame.

Without the spatial information, we can hardly distinguish the location of the entities. For example, the current temporal feature with the mean pooling over the spatial dimensions can reflect the object entities such as ``man" and ``woman". However, it can hardly tell if the man is close to the woman or far away from her when they are in the same frame. Thus, to generate better captions, we need to incorporate spatial information while still limiting the dimensions of the visual features to a level that is computationally feasible.

To overcome the above limitations, in this paper we introduce a novel video captioning model, where we
focus on the visual encoding part for generating a better video representation. In particular, we propose a simple spatial feature obtained by average pooling across different frame regions and integrate it into the captioning model. To tackle the issues with the noisy frames, we introduce a novel video refocusing encoder, where we encode each video twice. The first time is to select the key frame of a video by a key frame prediction network, and the second time is to re-encode the video centered at the key frame by two opposite directional RNN encoders. 

The main contributions of this work are threefold:
    \begin{itemize} 
    \item We propose a video refocusing encoder to predict the key frame and encode a video based on its key frame, and we integrate it within the captioning model. With no additional annotation, we further propose a novel reinforcement-learning based training method to jointly train the video refocusing encoder with the captioning model in an end-to-end manner. To our knowledge, this is the first work to introduce an iterative video encoding module into video captioning.

	\item We also propose a simple spatial feature and integrate it into the video captioning pipeline to provide more accurate entity information in different regions of a video.
	\item We test out method on two widely used benchmark video captioning datasets -- MSR-Video-to-Text (MSR-VTT) \cite{xu2016msr} and YouTube-to-Text (MSVD) \cite{chen2011collecting}, and we achieve results that rival state-of-the-art methods and even outperform them in most cases.

\end{itemize}

\section{Related Work}

Video captioning task has been a hot topic in recent years. Early video captioning methods~\cite{guadarrama2013youtube2text,krishnamoorthy2013generating,thomason2014integrating} view it as a template-matching problem, which generates a video description based on the (subject, verb, object) triplets predicted by the visual classifiers.
V2T~\cite{venugopalan2015sequence} first involves the sequence-to-sequence structure into video captioning. It encodes the video with a Long-Short Term Memory (LSTM) network~\cite{Hochreiter1997Long} and decodes it with an LSTM-based language model. 
Lorenzo ~etal~\cite{ABCDE} proposed a boundary-ware video encoding network which can discover and leverage the hierarchical structure of the video frames.
In~\cite{wang2018video}, Wang etal~ proposed a reconstruction framework which reconstructs the video features from the generated sentences. 
By making the reconstructed video features close to the original inputs, their method can enforce the generated caption to closely match the input video.

Several attempts have been made to introduce motion features and audio features to enhance the video captioning performance~\cite{shen2017weakly,wang2018watch}.
\cite{shen2017weakly} introduces the 3D ConvNets (C3D)~\cite{tran2015learning} network which is trained on the action recognition task into the video captioning problem.
Apart from the visual features, Wang~etal~\cite{wang2018watch} enhanced the video captioning with the audio information, where they extracted the audio information with VGGish~\cite{Hershey2016CNN} which was trained on the audio recognition task.
%
Inspired by the recent image captioning work~\cite{rennie2017self}, which uses Reinforcement Learning
(RL) to address the loss-evaluation mismatch or exposure bias problem
by including the inference process as a baseline in training, several attempts have been made to use reinforcement learning for the video captioning task~\cite{wang2018video, Pasunuru2017Reinforced, li2019end, chen2018less, phan2017consensus}.
To overcome the unstable of RL training, Pasunuru~etal~\cite{Pasunuru2017Reinforced} proposed a mixed loss which combines the cross-entropy with RL-based reward.
Xin~etal~\cite{wang2018video} divided the sentences into segments and decoded them with a hierarchical decoder trained with RL.

All these existing researches on video captioning mainly focus on sentence decoding, while the video encoding part is fairly standard. In contrast, we feel that the conventional one-step video encoding is insufficient to extract useful information. Thus, we propose a refocused video encoder by looking at a video twice to predict a better key-frame and extract better global and local visual features, which are optimized with a novel training loss.

\section{Method}

\begin{figure*}[ht]
	\centering
	\vspace{2mm}
	\includegraphics[width=\linewidth]{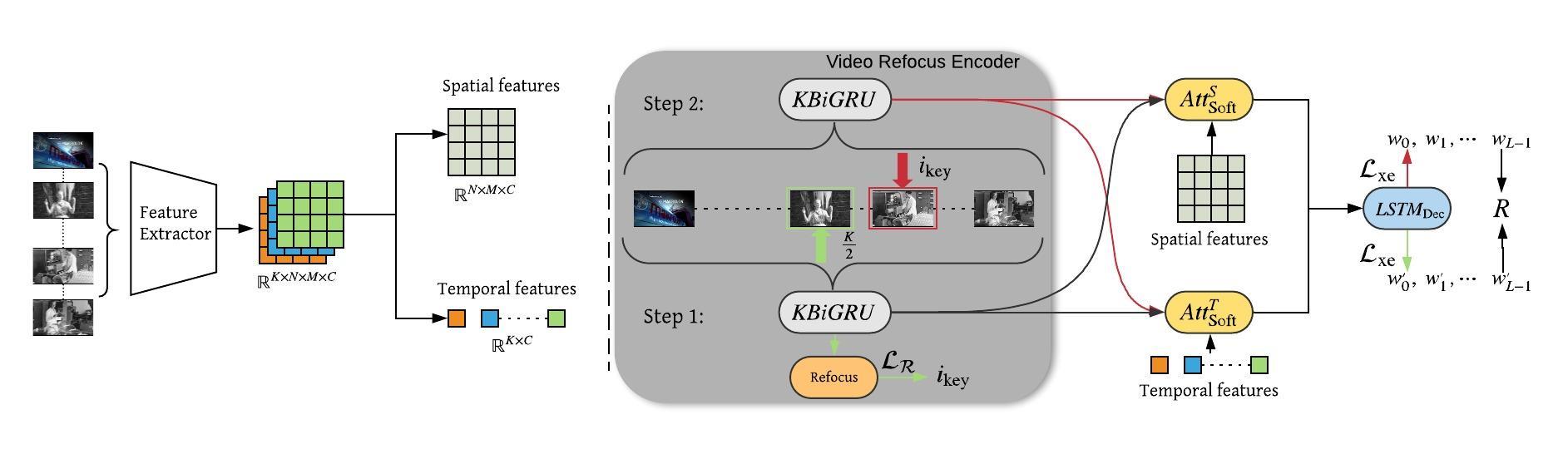} 
	\caption{Illustration of our proposed video captioning framework. Left: a CNN based video feature extractor to extract a spatial feature $v^s \in \mathbb{R}^{N\times M\times C}$ and a sequence of  
temporal features $\mathbf{V}^f \in \mathbb{R}^{K\times C}$ with $K$ being the number of video frames, $N\times M$ being the spatial resolution of each feature map, and $C$ being the number of feature channels. Right: the proposed refocus-based video captioning model, containing a Video Refocusing Encoder or VRE (in the grey-colored block), two soft-attention modules $\textit{Att}_{\text{Soft}}^S$ and $\textit{Att}_{\text{Soft}}^T$, and a LSTM-based sentence decoder $\textit{LSTM}_{\text{Dec}}$. The VRE consists of a key frame based Bidirectional GRU (\textit{KBiGRU}) to encode the temporal video features $\mathbf{V}^f$, and a refocusing module $F_{\text{Refocus}}$ to predict the key frame $i_{\text{key}}$.} \label{fig:framework}
\end{figure*}

Fig.~\ref{fig:framework} shows our overall video captioning framework. The left part shows a CNN-based video feature extractor, which extracts a spatial feature $v^s$ of dimensions $N\times M\times C$ and a sequence of
temporal features $\mathbf{V}^f$ of dimensions $K\times C$  with $K$ being the number of video frames, $N\times M$ being the spatial resolution of each feature map, and $C$ being the number of feature channels. 
The right part of the figure shows the proposed refocus-based captioning model. 
 It contains a Video Refocusing Encoder or VRE (in the grey-colored block), two soft-attention modules $\text{Att}_{\text{Soft}}^S$ and $\text{Att}_{\text{Soft}}^T$, and a LSTM-based sentence decoder $\text{LSTM}_{\text{Dec}}$. The {VRE} consists of a key frame based Bidirectional GRU ({KBiGRU}) to encode the temporal video features $\mathbf{V}^f$, and a refocusing module $F_{\text{Refocus}}$ to predict the key frame $i_{\text{key}}$. The two soft-attention modules are used to attend the spatial feature $v^s$ and the temporal features $\mathbf{V}^f$, respectively. The sentence decoder is a simple LSTM-based decoder~\cite{cho2014learning}.

Note that the {VRE} operates in two steps. In the first step, the model generates an initial caption using a  default key frame (the frame at the middle) and then predicts a new key frame by refocusing on the temporal video features with the refocusing module. In the second step, the model re-encodes the video features based on the refocused key frame $i_{\text{key}}$ and decodes a better caption. 
To train the refocusing module, we introduce a novel reinforcement learning based loss $\mathcal{L_R}$, and use it in combination with the cross-entropy loss $\mathcal{L}_{\text{xe}}$ to guide the key frame prediction.

\subsection{Video Feature Extractor}
\label{sec:VFE}

Given an input video $\mathcal{I}$, we first sampled the video frames $\{I_0, \cdots, \\ I_{K-1}\}$, where $K$ is the number of sampled frames in the video. We then extract the image feature for each frame $I_i$ using a pre-trained CNN model,

\begin{align}
v_i = \mathrm{CNN} (I_i)
\end{align}
where $v_i\in \mathbb{R}^{N \times M \times C}$ is the extracted feature from the final convolutional layer of the CNN with $N$ and $M$ being the height and width of each feature map, and $C$ being the number of channels.

After that, we apply the average pooling operations along the spatial and temporal dimensions to get the temporal and spatial features respectively as follows:

\begin{align}
v^f_i  = \frac{1}{N\times M}\sum_{n,m} v_{i,n,m} \hspace{2em} \text{for } i = 0, \ldots, K-1\\
v^s  = \frac{1}{K}\sum_{i} v_{i,:,:}
\end{align}
where $v^f_i\in \mathbb{R}^{C}$ is the temporal feature of $i$-th frame in the video, $v^s\in \mathbb{R}^{N\times M\times C}$ is the spatial feature of the entire video.

\subsection{Video Refocusing Encoder}

Previous work in video captioning  usually views a video as a sequence of temporal features, and uses RNNs to encode the features. Typical uni- and bi-directional RNNs such the  LSTMs have been used extensively in the video captioning task. These encoders always start from the first frame and end at the last one. However, if the duration of the event is not from the first to the last frame, the frames at the beginning and at the end may induce noise into the model. Since RNNs encode the features sequentially -- at each step updating the previous hidden state with information from the current input, the initial irrelevant frames can have a strong negative effect in  encoding the most relevant features for generating an informative caption. To circumvent this issue, we propose an encoding approach based on the idea of key frames.

Given the extracted temporal features $\mathbf{V}^f=[v^f_0,\cdots, v^f_{K-1}]$, and the index of a key frame $i_{\text{key}}$, we propose a novel Key frame based Bi-directional GRU (KBiGRU) to encode the temporal features:

\begin{equation}
    \overleftrightarrow{h} = \text{KBiGRU}(\mathbf{V}^f, i_{\text{key}}).
\end{equation}

\noindent In particular, we use two separate GRUs to encode the features: a right-to-left GRU ($\overleftarrow{\text{GRU}}$) to encode the features $v^f_{i_\text{key}}, \ldots, v^f_0$, and a left-to-right GRU ($\overrightarrow{\text{GRU}}$) to encode the features $v^f_{i_\text{key}}, \ldots, v^f_{K-1}$, as described by the following sequence of equations.

\begin{align}
\overleftarrow{h}_i  = \overleftarrow{\text{GRU}}(v^f_i,\overleftarrow{h}_{i+1}), \hspace{1em} \text{for } i \in [i_{\text{key}}, 0] \label{eq:backrnn}\\  
\overrightarrow{h}_j   = \overrightarrow{\text{GRU}}(v^f_j,\overrightarrow{h}_{j-1}), \hspace{1em} \text{for } j \in [i_{\text{key}},K-1] \label{eq:fwdrnn}\\
\overleftrightarrow{h}  = \tanh(\mathbf{W}_c  [\overleftarrow{h}_0, \overrightarrow{h}_{K-1}] +\mathbf{b}_c),
\end{align}

\noindent where $\mathbf{W}_c$ and $\mathbf{b}_c$ are the weight matrix and bias vector of the linear layer to combine the two encoder outputs and generate a final output. The hidden states of the GRUs (i.e., $\overleftarrow{h}_{i_{\text{key}+1}}$ in Eq. \ref{eq:backrnn} and $\overrightarrow{h}_{i_{\text{key}-1}}$ in Eq. \ref{eq:fwdrnn}) are initialized with zero vectors. In this way, the GRU encoders can get the main event accurately and encode starting from the most relevant frames. The main challenge, however, is how to select the key frame, especially since no explicit supervision about the key frame is given.

To obtain indirect information about the key frames, we propose to \emph{refocus} on the video frames. Specifically, our Video Refocusing Encoder (VRE) includes a KBiGRU and a refocusing module. Initially, we set the key frame index in the middle of the sequence, and encode an initial video feature through our KBiGRU encoder. This feature is then used to `refocus' on the temporal features and to select the (best) key frame in the frame sequence. With the updated key frame, we encode the video again and generate a `refocused' feature for the video. We concatenate the output of the \text{KBiGRU} in the two steps as the input to a linear layer to generate the overall video representation. The audio features are also adapted in the VRE to support additional global information of the video. The pseudocode of the process is shown in Algorithm~\ref{alg:VRE}. 
Note that an optional audio feature $v^a$ is used in the VRE to provide additional global information of a video.

\begin{algorithm}[t!]
\caption{Video Refocusing Encoder (VRE)}
\label{alg:VRE}
\begin{algorithmic}
\STATE{Input: temporal features $\mathbf{V}^f$ and audio feature $v^a$ (optional)}
\STATE {Output: video representations $o_1$ and $o_2$}
\STATE {Initialize $i_{\text{key}}$ with $K/2$}

/*\textbf{Step 1}*/
\STATE{$\overleftrightarrow{h^1} =\text{KBiGRU}(\mathbf{V}^f, i_{\text{key}})$}
\STATE{$o_1 = \tanh(\mathbf{W}_a[\overleftrightarrow{h^1}, v^a] + \mathbf{b}_a)$}
\STATE{$i_{\text{key}} = F_{\text{Refocus}}(o_1, \mathbf{V}^f)$}

/*\textbf{Step 2}*/
\STATE{$\overleftrightarrow{h^2} = \text{KBiGRU}(\mathbf{V}^f, i_{\text{key}})$}
\STATE{$\tilde{o}_2 =\tanh(\mathbf{W}_m [\overleftrightarrow{h^1}, \overleftrightarrow{h^2}] +\mathbf{b}_m)$}
\STATE{$o_2 = \tanh(\mathbf{W}_a[\tilde{o}_2, v^a] + \mathbf{b}_a)$}
\end{algorithmic}
\end{algorithm}

\subsubsection*{\textbf{Refocusing Module}}
 
The refocusing module uses soft attentions to identify the key frame. It uses the initial video representation $o_1$ (see Alg. \ref{alg:VRE}) as the query vector and computes attention weights for the temporal video features in $\mathbf{V}^f$. More formally,

\begin{align}
    \label{11}
   a_i  = \mathbf{W}_{\alpha}\tanh(\mathbf{W}_o o_1 \otimes 
   \mathbf{W}_v v_i^f) + {b}_{\alpha}\\
   \boldsymbol{\alpha}  = \text{softmax}(\mathbf{a})
\end{align}

\noindent where $\otimes$ represents element-wise multiplication, $\mathbf{W}_o$ and $\mathbf{W}_v$ are the mapping weights to transform $o_1$ and $v_i^f$, respectively, and $\mathbf{W}_{\alpha}$ and ${b}_{\alpha}$ are the weight and bias of a linear layer to transform the input to a real value. The attention weights $\boldsymbol{\alpha} = (\alpha_0,\ldots, \alpha_i, \ldots, \alpha_{K-1})$ represent how much each frame is relevant with respect to the main event. We select the frame with the largest $\alpha$ as the key frame, $i_{\text{key}}$.

\begin{align}
   i_{\text{key}}   = F_{\text{Refocus}}(o_1, \mathbf{V}^f) \\
                      = \arg\max_i(\{\alpha_0,\ldots, \alpha_i, \ldots, \alpha_{K-1}\}) \label{eq:keyframe}
\end{align}

\subsection{Sentence Decoder}

Our LSTM-based sentence decoder uses two attention modules along with the video representation from the VRE module. 

\subsubsection*{\textbf{Attention Modules}}

As shown in Fig. \ref{fig:framework}, we use two soft attention modules to provide contextual video information to the sentence generator. One module attends to the spatial features while the other attends to the temporal features as follows.

\begin{align}
    v_{\text{Att}}^S= \textit{Att}_{\text{Soft}}^{S}(v^s, h_t)\\
    v_{\text{Att}}^T= \textit{Att}_{\text{Soft}}^{T}(\mathbf{V}^f, h_t)
\end{align}
where $h_t$ is usually the decoder hidden state at time $t$. More formally, the temporal attention module  $\textit{Att}_{\text{Soft}}^T$ is defined as:

\begin{align}
    \label{11}
   a_i^T  = \mathbf{W}_{\beta}\tanh(\mathbf{W}_h h_t \otimes 
   \mathbf{W}_v^{\beta} v_i^f) + {b}_{\beta}\\
   \label{12}
   \boldsymbol{\alpha}^T  = \text{softmax}(\mathbf{a}^T)\\
   \label{13}
   v_{\text{Att}}^T =\sum_{i=0}^{K-1}\alpha_i^T v^f_i
\end{align}
where $\mathbf{W}_h$, $\mathbf{W}_v^{\beta}$, $\mathbf{W}_{\beta}$ and ${b}_{\beta}$  are similarly defined as before. The spatial attention module $\textit{Att}_{\text{Soft}}^{S}$ has the same structure as $\textit{Att}_{\text{Soft}}^{T}$, but attends over the $N \times M$ spatial regions, generating $N \times M$ different attention weights for each decoder state.

\subsubsection*{\textbf{Overall Framework}}

Algorithm~\ref{alg:Overall} describes our overall decoding process. First, the VRE generates the two video representations $o_1$  and $o_2$ using Alg. \ref{alg:VRE}. The video representations of each step is concatenated with the (previous) hidden state of the decoder LSTM, which are then used for temporal and spatial attentions in $\textit{Att}_{\text{Soft}}^T$ and $\textit{Att}_{\text{Soft}}^S$ to generate the respective context vectors, $v_{\text{Att}}^T$ and $v_{\text{Att}}^S$. Then, the decoder LSTM generates its current hidden state with the concatenated features of each step, which is in turn used for generating the next word (not shown in Alg. \ref{alg:Overall}). We use the $\mathbf{0}$ vector and the video representations $o_i$ to initialize the hidden state $h^i_{-1}$ and memory cell $c^i_{-1}$ of the LSTM decoder. The green and red lines in Fig.~\ref{fig:framework} represent the first and the second steps, respectively, while the black lines represent the processes that exist in both steps. Note that the video representations are also used as an input to the LSTM decoder at $t=0$, which is later replaced by the word embedding of the previous word $W_{y_{t-1}}^{emb}$.

\begin{algorithm}[t!]
\caption{Overall framework}
\label{alg:Overall}
\begin{algorithmic}
\STATE {Input: visual features $\mathbf{V}^f$, $v^s$ and audio feature $v^a$}
\STATE {Output: hidden states $\{h^1_0$, $h^1_1$, $h^1_2$, ..., $h^1_L\}$ }
\STATE{\ \ \ \ \ \ \ \ \ \ \ \ \ \ and $\{h^2_0$, $h^2_1$, $h^2_2$, ..., $h^2_L\}$   for 2 steps}
\STATE {$o_1, o_2 = \text{VRE}(\mathbf{V}^f, v^a)$}
\FOR {$i = 1 \to 2$}
\STATE {Initialize $h^{i}_{-1}$ with $\mathbf{0}$}
\STATE {Initialize $c^{i}_{-1}$ with $o_i$}
\FOR {$t = 0 \to L - 1 $}
\STATE {$v_{\text{Att}}^T = \textit{Att}^T_{\text{Soft}}(\mathbf{V}^f,[o_i, h^{i}_{t-1}])$}
\STATE {$v_{\text{Att}}^S = \textit{Att}^S_{\text{Soft}}(v^s, [o_i, h^{i}_{t-1}])$}
\IF {t = 0}
\STATE {$h^{i}_t, c^{i}_t = LSTM([v_{\text{Att}}^T, v_{\text{Att}}^S, v^a, o_i],
h^{i}_{t-1}, c^{i}_{t-1})$}
\ELSE
\STATE {$h^{i}_t, c^{i}_t = LSTM([v_{\text{Att}}^T, v_{\text{Att}}^S, v^a, W^{emb}_{y_{t-1}}],
h^{i}_{t-1}, c^{i}_{t-1})$}
\ENDIF
\ENDFOR
\ENDFOR
\end{algorithmic}
\end{algorithm}

\subsection{Training with Cross Entropy and Reinforcement Learning Losses}

\textbf{Cross-Entropy Loss.} Let $Y$ denote the ground-truth caption. Following previous work, we use the cross-entropy loss for the two steps, as defined below.

\begin{align}
    \mathcal{L}_{\text{xe}}^{\text{STEP1},e}=  -\sum_{t}\log p^{\text{STEP1}, e}(y_t|y_{1:t-1},\mathbf{V}^f,v^s, v^a) \label{eq:step1xe}\\
    \mathcal{L}_{\text{xe}}^{\text{STEP2},e}=  -\sum_{t}\log p^{\text{STEP2},e}(y_t|y_{1:t-1},\mathbf{V}^f,v^s, v^a) \label{eq:step2xe}
\end{align}
where $p^{\text{STEP1},e}(y_t|.)$ and $p^{\text{STEP2},e}(y_t|.)$ are the probabilities of a ground-truth word $y_t$ in the two steps at training epoch $e$.

\noindent\textbf{Reinforcement Learning Loss.} For the refocusing module, since we do not have explicit labels for the key frames, we propose to use the predicted caption-level probabilities to guide the optimization. The log probabilities of a caption in the two steps can be achieved by simply negating the cross-entropy losses in Eq. \ref{eq:step1xe}-\ref{eq:step2xe} for the caption, i.e., $- \mathcal{L}_{\text{xe}}^{\text{STEP1},e}$ and $- \mathcal{L}_{\text{xe}}^{\text{STEP2},e}$. These probability scores reflect how good the model has been fitted to generate the respective caption for a video. Higher score means that the model prefers to generate the ground-truth caption rather than other candidate sentences.

In our model, the VRE provides two video representations, $o_1$ and $o_2$. Ideally, the refocused representation $o_2$ should contain more relevant video information than the initial  representation $o_1$ since $o_2$ is generated from both the default and refocused key frames (see $\overleftrightarrow{h^1}$ and $\overleftrightarrow{h^2}$ in Alg. \ref{alg:VRE}), whereas $o_1$ is generated only from the default key frame. Therefore, for a given video, the sentence probability generated with $o_1$ should be lower than the probability generated with  $o_2$, and this increase in the predicted probability scores can be considered as the contribution from the refocused key frame selection and the refined encoding in the second step. The larger the difference, the more representative the key frame is. Thus, it can be used to optimize the selection of the refocused key frame as a more direct loss in addition to the (indirect) cross-entropy loss.

Since this loss is computed only after the full caption is generated, we formulate this as a reinforcement learning loss as follows. 
\begin{align}
    \mathcal{L_R} =  - (R_{e} - R_{e-1}) \log(\alpha_{i_{\text{key}},e}) \label{eq:rl}\\
    R_{e}=  - (\mathcal{L}_{\text{xe}}^{\text{STEP2},e}-\mathcal{L}_{\text{xe}}^{\text{STEP1},e})
\end{align}
where $R_{e}$ is the reward for the refocusing module in training epoch $e$, and $\alpha_{i_{\text{key}},e}$ is the probability of the key frame (Eq. \ref{eq:keyframe}) in epoch $e$. 

In the loss, $R_{e-1}$ acts as a moving threshold to encourage the refocused key frame to be more representative from one epoch to the next. The model encourages the current key frame only when the reward of the sentence in the current epoch is larger than the reward in the previous epoch. When $R_{e} > R_{e-1}$, this means that the current $o_2$ (and the key frame) is more useful to the decoder in generating a better caption compared with the key frame selected in the previous epoch. In this situation, $\mathcal{L_R}$ is positive,  forcing probability of $\alpha_{i_{\text{key}}}$ to increase. When $R_{e} < R_{e-1}$, it means that $o_2$ (and the key frame)  is worse than the one in the previous epoch, thus the current key frame should be changed. In this situation, the loss is a negative one, which forces $\alpha_{i_{\text{key}}}$ to decrease.

In practice, the derivative of the softmax function tends to be close to 0 when $\alpha_{i_{\text{key}}}$ is close to 1. This does not generate enough gradient in later epochs if  $\alpha_{i_{\text{key}}}$ is already close to 1, causing the model fail to select the key frame.  To overcome this problem, we modify the loss function as follows.

\begin{align}
    \mathcal{L_R}^+ = - (R_{e} - R_{e-1})(\mathbbm{1}^+log(\alpha_{i_{\text{key}},e}) + \mathbbm{1}^-log(1 - \alpha_{i_{\text{key}},e})),    
\end{align}
where $\mathbbm{1}^+$ and $\mathbbm{1}^-$ are indicator functions defined as follows. If $R_{e} > R_{e-1}$, then $\mathbbm{1}^+ = 1$ else $\mathbbm{1}^+ = 0$; similarly, $\mathbbm{1}^- = 1$ when $R_{e} < R_{e-1}$ and $0$ otherwise. When $R_{e} > R_{e-1}$, the loss is the same as in Eq. \ref{eq:rl}. However, when $R_{e} < R_{e-1}$ and $\alpha_{i_{\text{key}}}$ is close to 1, the modified loss of $log(1 - \alpha_{i_{\text{key}}})$ will generate enough gradient to decrease $\alpha_{i_{\text{key}}}$.

\noindent\textbf{Combined Loss.} Our overall training loss includes both the cross-entropy loss and our reinforcement learning loss. Initially, we pre-train the model with the cross-entropy loss for first few epochs to gain a reliable model. Then, we train the refocusing module and captioning module jointly with the following combined loss, 
\begin{align}
\label{23}
\mathcal{L}_{\text{All}}  = \mathcal{L}_{\text{xe}}^{\text{STEP2},e}+\mathcal{L}_{\text{xe}}^{\text{STEP1},e} + \beta\mathcal{L_R}^+,    
\end{align}
where $\beta$ is a weight parameter to control the contribution of the reinforcement learning loss. In our model, we set $\beta$ to 0.03.

\section{Experiment}

\subsection{Datasets and Metrics}
\textbf{MSR-Video to Text (MSR-VTT) \cite{xu2016msr}.} MSR-VTT is a large-scale video description dataset which contains 10,000 video clips collected from the Internet. The videos contain audio resource and last longer than 14.8s on average. For each video, it is annotated with about 20 natural sentences. We separate the dataset into three contiguous parts contains 6,512, 498, 2,990 for training, validation, and testing.

\textbf{Microsoft Video Description Corpus (MSVD) \cite{chen2011collecting}.} MSVD consists of 1,970 Youtube video clips, and 85K English descriptions collected by Amazon Mechanical Turkers. The audio sources are muted in the video clips by the provider.
Following the previous works~\cite{guadarrama2013youtube2text,venugopalan2014translating}, we split the dataset into a 1,200 training set, 100 validation set, and 670 testing set by the contiguous index number. 

\textbf{Evaluation Metrics}
We use four widely-used metrics in the test process:
BLEU~\cite{papineni2002bleu}, METEOR~\cite{denkowski2014meteor}, CIDEr~\cite{Vedantam2015CIDEr}, and ROUGE$\_$L~\cite{lin2004rouge}.
 BLEU metrics calculate the matching level of the n-grams in candidate and reference captions. In the test, we applied BLEU4, which contains 4 words in each valid sub-sequence, as one of the evaluation metrics.

METEOR evaluates the captions by aligning candidate n-grams with multiple ground-truth in different forms like stem, synonym, and paraphrase to generate a better semantical match.
CIDEr calculates cosine similarity on n-gram word sub-sequences of the candidate captions and references and performing a Term Frequency-Inverse Document Frequency among the test dataset.
ROUGE$\_$L not only does the n-gram matching but also consider the longest common sub-sequence (LCS) which is also significant for the language evaluation. 
\subsection{Training Details}

We apply ResNet152~\cite{he2016deep} trained on the ImageNet dataset~\cite{krizhevsky2012imagenet} as pre-trained CNN network as our feature extractor without fine-tuning.  For each video, we uniformly extract 20 frames. The raw features are the concatenation of frame features from the last layer of the CNN network with a size of $7 \times 7 \times 2,048$. Then we apply the temporal and spatial average pooling introduced in Section~\ref{sec:VFE} on the raw video features to generate temporal and spatial features. Inspired by the success of audio-based features in~\cite{wang2018watch}, we also extract audio features using VGGish~\cite{Hershey2016CNN}, and the audio feature is a 128-dim vector for each video. To extract audio features, we set the sample rate as 44,100 and the bit rate as 160 kbps in our model. Note that we do not use audio features $v^a$ for MSVD dataset since there is no audio source.

Before training, we build a vocabulary for the corpus. We make each letter lower-case, filter the words that appear less than three times and replace them with the signal ``$<$UNK$>$". We set the maximum length of captions to be 15. If a ground-truth has more than 15 words, we abandon the words after the 15-th words. At the end of a caption, we add a signal ``$<$EOS$>$" to represent the end of the caption. If the length of a sentence is shorter than the 15, we add ``$<$PAD$>$"s after the ``$<$EOS$>$" to keep the length of the sentence 15.

In our captioning model, we set the size of the hidden state 512 for both the encoding and decoding parts. The size of the word embedding is also 512. We adopt a dropout-out layer with a dropout-out rate of 0.5 at the end of the LSTM decoder. 
In the training process, we set the learning rate $3e^{-4}$ and the scheduled sampling rate in the decoding process 0.25. We first train the decoder and STEP1 of the VRE in Alg.~\ref{alg:VRE} for 5 epochs, to gain a baseline that makes the refocus module easier to be optimized. At the first 5 epochs, we only apply cross-entropy loss $\mathcal{L}_{\text{xe}}$ on the predicted captions in STEP1. After that, we activate the video refocusing module and train the model with the combined loss. Note that the cross-entropy loss in the combined loss contains not only the loss of captions generated in STEP1 but also that of STEP2. We also apply reinforcement learning for the decoder with CIDEr and ROUGE$\_$L as rewards to fine tune the model.

In the experiment, we find that the model with constant key frame inputs can be optimized better than the model that trains the captioning and refocusing modules at the same time. This is because, when the key frames are varying, the encoder has to accommodate different key frames to ensure the captioning model can still generate some meaningful sentences, which makes the model incapable of exploring detailed features of different key frames. Thus, we first train the captioning model jointly with the refocusing model. Then, we pick out the key frames predicted in the best model and save them as constants. Finally, we train the model without the refocusing module but using the saved key frames as the input.

\subsection{Results}

\textbf{Results on MSR-VTT and Ablation Study.} Table~\ref{tab:msr-vtt} shows the results of different methods on the MSR-VTT dataset. We compare our method with MeanPool~\cite{venugopalan2014translating} and SoftAtt~\cite{Yao_2015_ICCV} trained with only temporal features to shown the contribution of the proposed spatial feature. We also compare with the recent works HRL~\cite{wang2018video}, $M^3$~\cite{wang2018m3}, RecNet~\cite{wang2018reconstruction}, MFA~\cite{long2018video}, $\mathbf{FSTA}$\cite{liu2018fine} and PickNet~\cite{chen2018less}, which have achieved outstanding performance.

To analyze the contribution of each component in our model, we also consider the following baselines. 1) $\mathbf{V}^f + v^s$ model: It only contains the two soft-attention modules and the LSTM sentence decoder, without any RNN encoder. The soft-attention modules are guided by the hidden state of the LSTM decoder at each time step. 2) $\textbf{KBiGRU}(\mathbf{V}^f)$  model: It uses one-step \textit{KBiGRU} with the input of temporal features. The \textit{KBiGRU} encoded feature is then used to soft-attend the raw temporal features as well as being an initial input to the LSTM decoder. 3) $\textbf{KBiGRU}(\mathbf{V}^f)+v^s+v^a$ mode: It makes use of the additional spatial and audio features. 4) $\textbf{VRE}(\mathbf{V}^f+v^s+v^a)$: This is our full model. It is built on top of $\textbf{KBiGRU}(\mathbf{V}^f)+v^s+v^a$ model by applying the video refocusing module in the encoder, which is trained with the combined loss introduced above.  

\begin{figure*}[ht]
    \centering
    \includegraphics[width=\linewidth]{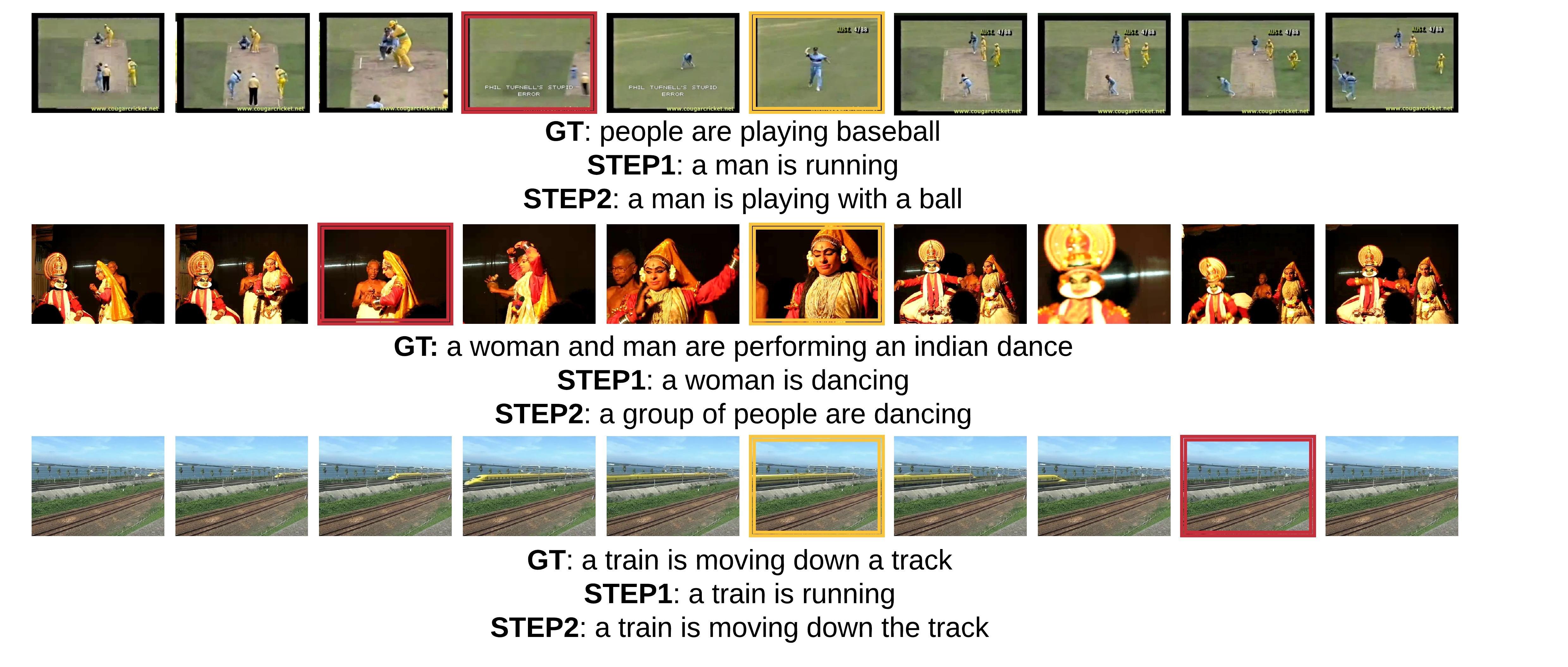} 
    \caption{Visualization results for the video refocusing module. Here  we only show the even frames to save space. The frames with yellow borders are 10-th frames used in STEP1 as the default key frame. The frames with red borders are the predicted key frame used in STEP2. \textbf{GT}: ground-truth captions; \textbf{STEP1}: generated captions in STEP1; \textbf{STEP2}: generated captions in STEP2.}
    \label{fig:visual_exa}
\end{figure*}
From Table~\ref{tab:msr-vtt}, we can see that, with the help of the spatial features, $\mathbf{V}^f + v^s$ model achieves better performance than the Mean-pooling and Soft-Attention models in all the four evaluated metrics. This is because the spatial features provide more detail information about the objects, thus resulting in more accurate and detailed captions compared with those generated by Mean-pooling and Soft-Attention models. $\textbf{KBiGRU}(\mathbf{V}^f)$ also achieves better results than the soft-attention method. It improves $\textbf{SoftAtt}$ by 9.5, 2.4 and 6.7 on BLEU4, METEOR and CIDEr, respectively. It shows that involving the \textit{KBiGRU} that gives a better global view of each video can lead to better performance. Comparing among the four variants of our method, the full model $\textbf{VRE}$ achieves the best performance in all the metrics except MEREOR, which demonstrates the effectiveness of individual components including the spatial features, the \textit{KBiGRU} module and the refocusing module with its associated combined loss. The performance of our full model is comparable to the state-of-the-art method HRL~\cite{wang2018video}, where we achieve better performance in BLEU4, ROUGE$\_$L, and CIDEr by 1.9, 0.3, 0.3, respectively. Note that our method is a complement to HRL~\cite{wang2018video}, in the sense that HRL introduces a better hierarchical sentence decoder while we introduce a better video encoder.

\begin{figure*}[ht]
    \centering
    \includegraphics[width=\linewidth]{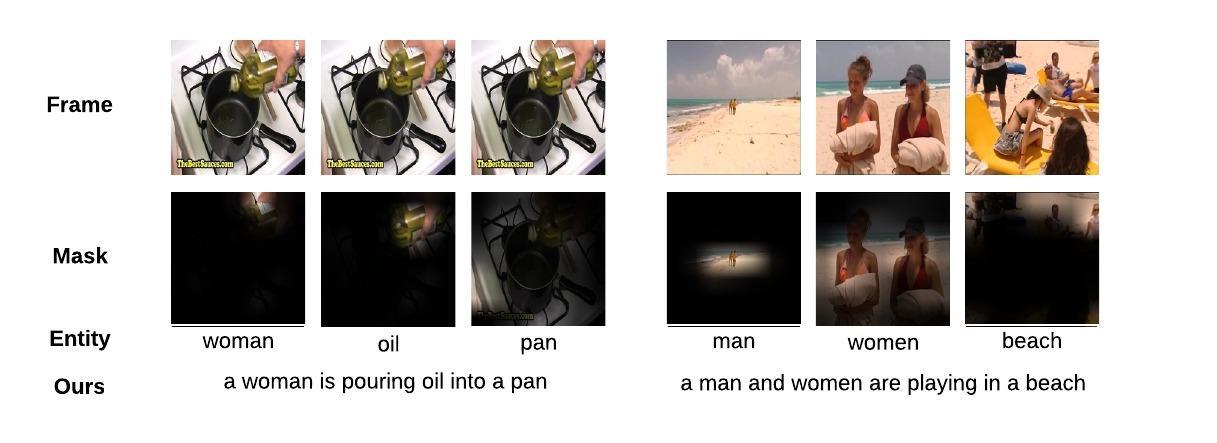} 
    \caption{Visualization results for the spatial feature. \textbf{Ours} are the generated captions. \textbf{Entities} are the object entities in the captions. 
    \textbf{Mask} is the visualized result generated by multiplying the spatial attention mask with a frame to show that the spatial attention is attending to each corresponding object entity.}
    \label{fig:mask}
\end{figure*}
 
\begin{table}
    \begin{center}
        \small
        \caption{Comparisons of the video captioning results of different methods on the MSR-VTT dataset.}
        \label{tab:msr-vtt}
        
        \begin{tabular}{l|c|c|c|c}
        
            \hline
            Model    & BLEU4 & ROUGE    & METEOR         & CIDEr \\
            \hline
            $\mathbf{MeanPool}$\cite{venugopalan2014translating} &30.4 &  52.0 &23.7 &35.0 \\ 
            $\mathbf{SoftAtt}$\cite{Yao_2015_ICCV} &28.5 & 53.3 &25.0 &37.1 \\
            \hline
            $\mathbf{RecNet_{local}}$\cite{wang2018reconstruction} &39.1 & 59.3 &26.6 &42.7 \\
            $\mathbf{HRL}$\cite{wang2018video} & 41.3  & 61.7 & $\mathbf{28.7}$ & $48.0$  \\
            $\mathbf{MFA}$\cite{long2018video} & 39.1 & -  & 26.7 & -  \\
            $\mathbf{PickNet}$\cite{chen2018less}   & 41.3 & - & 27.7 & 44.1  \\
            $\mathbf{M^3-VC}$\cite{wang2018m3} &38.1 & - &26.6 &- \\
            $\mathbf{FSTA}$\cite{liu2018fine} & 39.8 & - & 26.5 &41.1 \\

            \hline
            $\mathbf{V}^f + v^s$    & 34.5 & 55.3 & 26.7         &39.6\\
            $\textbf{KBiGRU}(\mathbf{V}^f)$     & 38.0 & 58.0 & 27.4         &43.8\\
            
            $\textbf{KBiGRU}(\mathbf{V}^f)+v^s+v^a$    & 40.7 & 60.4 & 27.8 &45.1\\
            $\textbf{VRE}(\mathbf{V}^f,v^s,v^a)$    & $\mathbf{43.2}$ & $\mathbf{62.0}$ & 28.0 & $\mathbf{48.3}$         \\
            \hline
        \end{tabular}
    \end{center}
\end{table}

\begin{table}
    \begin{center}
        \small
        \caption{Comparisons of the video captioning results of different methods on the MSVD dataset.}
        \label{tab:MSVD}
        
        \begin{tabular}{l|c|c|c|c} 
        
            \hline
            Model   & BLEU4 & ROUGE     & METEOR         & CIDEr \\
            \hline
            $\mathbf{RecNet_{local}}$\cite{wang2018reconstruction} & $\mathbf{52.3}$ &69.8 & $34.1$ &80.3 \\ 
            $\mathbf{MFA}$\cite{long2018video}    & 52.0 & -    & 33.5    & 72.0  \\
            $\mathbf{PickNet}$\cite{chen2018less}   & $\mathbf{52.3}$ & 69.6    & 33.3    & 76.5  \\
            $\mathbf{TubeFeat}$\cite{zhao2018video} & 43.8 &69.3 &32.6 &52.2 \\
            $\mathbf{TSAED}$\cite{wu2018interpretable} & 51.7 & - & 34.0 &74.9 \\
            $\mathbf{FSTA}$\cite{liu2018fine} & 51.2 & - & 32.5 &70.6 \\
            \hline
            $\textbf{VRE}(\mathbf{V}^f,v^s)$    & 51.7 & $\mathbf{71.9}$    & $\mathbf{34.3}$         & $\mathbf{86.7}$         \\
            \hline
        \end{tabular}
    \end{center}
\end{table}

\begin{table}
    \begin{center}
        \small
        \caption{Comparisons of our method and the multitask video captioning models.}
        \label{tab:multi}
        
        \begin{tabular}{l|c|c|c|c} 
        
            \hline
            Model  & BLEU4 & ROUGE     & METEOR         & CIDEr \\
            \hline
            MSR-VTT \\
            \hline
            $\mathbf{SibNet}$\cite{liu2018sibnet} & 40.9 & 60.2 & 27.5 & 47.5\\
            $\mathbf{E2E}$\cite{li2019end} & 40.4 & 61.0 & 27.0 & $\mathbf{48.3}$\\
            \hline
            $\textbf{VRE}(\mathbf{V}^f,v^s,v^a)$  & $\mathbf{43.2}$  & $\mathbf{62.0}$    & $\mathbf{28.0}$         & $\mathbf{48.3}$         \\
            \hline
            MSVD \\
            \hline
            $\mathbf{SibNet}$\cite{liu2018sibnet} & $\mathbf{54.2}$ & $71.7$ & $\mathbf{34.8}$ & $\mathbf{88.2}$\\
            $\mathbf{E2E}$\cite{li2019end} & 50.3 &70.5 &33.6 &86.5\\
            \hline
            $\textbf{VRE}(\mathbf{V}^f,v^s)$    & 51.7 & $\mathbf{71.9}$    & 34.3         & $86.7$         \\
            \hline
        \end{tabular}
    \end{center}
\end{table}

\textbf{Results on MSVD.} Table~\ref{tab:MSVD} shows the results of different methods on MSVD dataset. Since MSVD dataset has no audio, we do not apply audio feature in the model. Compared with the recent video captioning methods that report their results on MSVD, our method achieves the best performance in CIDEr and ROUGE$\_$L. Our CIDEr score is significantly higher than the previous best score obtained by $\mathbf{RecNet_{local}}$, by a large margin of 6.4. ROUGE$\_$L and METEOR of our model are also better than those of $\mathbf{RecNet_{local}}$, achieving better performance by 2.1 and 0.2, separately.

\textbf{Comparisons with Multitask Models.} Recently, there are some video captioning works coupling with some other task to get more suitable video representations. In particular, in \textbf{E2E}~\cite{li2019end}, Li et al. trained the captioning model together with an attribute prediction model. $\mathbf{SibNet}$~\cite{liu2018sibnet} trains two individual branches that involve an auto-encoder network and SAN~\cite{Lin2017A}, respectively. Table~\ref{tab:multi} shows the corresponding results. It can be seen that our single-task model is comparable with the two multitask models, achieving the best performance in the large-scale MSR-VTT dataset in all the four matrices, BLUE4, ROUGE$\_$L and METEOR, and CIDEr. And our model also achieves the best ROUGE$\_$L score in the middle-scale MSVD dataset.

\textbf{Qualitative Results.} 
Fig.~\ref{fig:visual_exa} shows the visualized results for the key frame selection. We can see that in the first example, the video consists of multiple noncontagious scenarios. The default key frame in STEP1 locates in the sub-sequence that a man is running for the ball. Captioning based on this key frame is ``a man is running", which only contain the information of the frame sub-sequence and lead to a misunderstanding of the video. In contrast, our model predicts the 6-th frame (the fourth one in the example) as the key frame, which is the junction of ``the man is hitting the ball" and `` the man is running". Thus, the model can fully encode the entire event and connect the actions ``running" with ``hitting" to predict a better caption shown in \textbf{STEP2}. In the second example, our model views the breakout point of the dance as the key frame so as to separate the video into two parts which all represent the same event of multiple people dancing and can be well encoded by the \textit{KBiGRU} module. 
This is clearly better the default key frame that focuses on the single woman and predicts the caption according to it. In the experiments, we find that if the default key frame in STEP1 is good enough to represent the entire event, the predicted key frame tends to be located at the beginning or the end of the event so as to learn a better representation in single temporal order. An example is shown in the third case in Fig.~\ref{fig:visual_exa}, where the predicted key frame is located at the end of the event.

Fig.~\ref{fig:mask} shows the visualized results of the spatial attention, where brighter regions in the masks refer to the regions with higher spatial attention weights. 
In the left example, we can see that our attention model can accurately focus on the ``hand", ``oil" and ``pan", when respectively predicting the ``woman", ``oil" and ``pan" entities in the sentence. In the right example, where there are multiple scenarios in the video, the spatial attention mask can still focus on different entities in different frames. 
These results indicate that the introduced spatial features obtained by temporal mean-pooling of different spatial blocks can represent different entities in a video. 
\section{Conclusion}

In this paper, we have introduced a novel video refocusing encoder and a simple spatial feature for video captioning task. Video refocusing encoder uses a refocusing attention module to predict the key frame according to the video content and then re-encode the video according to the key frame to get a more representative video feature. The spatial feature is simply generated from the CNN feature extraction network with the mean pooling across the temporal dimension. The experimental results on MSR-VTT and MSVD dataset have demonstrated the effectiveness of the proposed method.

\bibliographystyle{unsrt}  

\bibliography{sample-base}

\end{document}